\documentclass{article}

\PassOptionsToPackage{numbers, compress}{natbib}

\usepackage[final]{./style/neurips_2023}

\usepackage[utf8]{inputenc} 
\usepackage[T1]{fontenc}    
\usepackage{hyperref}       
\usepackage{url}            
\usepackage{booktabs}       
\usepackage{amsfonts}       
\usepackage{nicefrac}       
\usepackage{microtype}      
\usepackage{xcolor}         
\usepackage{lmodern}
\usepackage{amsmath}
\usepackage{graphicx}
\usepackage{adjustbox}

\newcommand{\approach}{\textsc{LLM-DTE} }

\title{Explaining Tree Model Decisions in Natural Language for Network Intrusion Detection}

\author{Noah Ziems, Gang Liu, John Flanagan, Meng Jiang \\
  University of Notre Dame \\
  \texttt{\{nziems2, gliu7, jflanag5, mjiang2\}@nd.edu}
}

\begin{document}

\maketitle

\begin{abstract}
  Network intrusion detection (NID) systems which leverage machine learning have been shown to have strong performance in practice when used to detect malicious network traffic.
  Decision trees in particular offer a strong balance between performance and simplicity, but require users of NID systems to have background knowledge in machine learning to interpret.
  In addition, they are unable to provide additional outside information as to why certain features may be important for classification.

  In this work, we explore the use of large language models (LLMs) to provide explanations and additional background knowledge for decision tree NID systems.
  Further, we introduce a new human evaluation framework for decision tree explanations, which leverages automatically generated quiz questions that measure human evaluators' understanding of decision tree inference.
  Finally, we show LLM generated decision tree explanations correlate highly with human ratings of readability, quality, and use of background knowledge while simultaneously providing better understanding of decision boundaries.

\end{abstract}

\section{Introduction}
\label{sec:introduction}

Network intrusion detection (NID) systems monitor incoming and outgoing network traffic to detect potentially malicious activity.
When suspicious traffic is detected, a system administrator is alerted for further investigation.
Early NID systems gained popularity by analysing traffic with simple rule-based approaches \cite{mukherjee1994network}.
However, over time new types of attacks such as malware, phishing, worms, sql injection, scan attacks, and distributed denial-of-service (DDoS) attacks have significantly increased the vulnerable attack surface of networks.
As the kinds of attacks increase, so too does the complexity of detecting them using rule-based methods alone.
In response, machine learning (ML) and data mining based NID systems which learn to detect malicious traffic from large amounts of real data began to emerge \cite{lee1999data}.
However, due to their black-box nature, understanding \textit{why} and \textit{how} a model has come to a particular conclusion remains a significant limitation of ML based NID systems in real-world applications.
While decision trees offer a strong balance between simplicity and performance, they still require a user familiar with their underlying mechanics to interpret why a particular prediction was made.
Efforts to assist in explaining decision-tree based NID systems often rely on providing additional information such as quantifying the importance of features for a given prediction \cite{guo2018lemna, barnard2022robust, wang2020explainable}.
How one can create decision tree explanations for NID systems that can be easily understood by users with no background knowledge of machine learning remains an open area of research.
Further, how to measure the quality of these explanations also remains an open question.

The recent success of large language models (LLMs) has enabled high quality open-ended text generation for a wide variety of tasks such as long-form question answering, summarization, arithmetic reasoning, and more.
In this work, we explore whether large language models can be used to generate high quality explanations from decision tree inference, and we explore new approaches for measuring the quality of decision tree explanations.

Our main contributions are as follows:
\begin{enumerate}
  \item We introduce Large Language Model Decision Tree Explanations (LLM-DTEs), an approach for generating decision tree explanations targeted at users with no background knowledge of machine learning.
  \item We introduce a new framework of human evaluation of decision tree explanations which relies on an evaluator's ability to answer counterfactual quiz questions.
  \item We apply \approach to network intrusion detection (NID) and show it scores highly with human evaluators.
\end{enumerate}

\begin{figure}[t]
    \centering
    \includegraphics[width=0.7\textwidth]{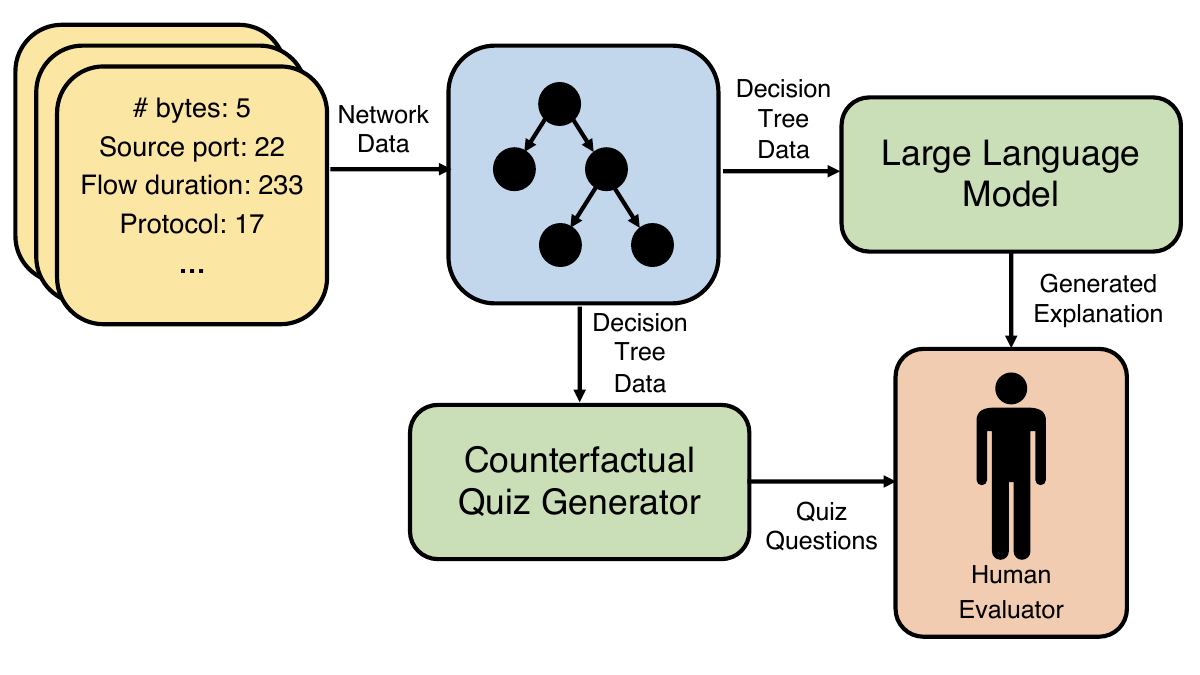}
    \vspace{-0.2in}
    \caption{The overall framework for our proposed approach (LLM-DTE). Given network traffic data, a decision tree evaluates the data and classifies it as either \textit{Benign} or \textit{Threat}. Path and structure data from the decision tree are then provided to a large language model to generate an explanation. At the same time, quiz questions are generated using rule-based templates (not LLMs). Finally, a human evaluator is provided with the generated explanation and tasked with answering quiz questions to test their understanding.}
    \label{fig:url-reconstructions}
    \vspace{-0.1in}
\end{figure}
\section{Related Work}
\label{sec:related_work}

\subsection{AI Explainability}
Recent years have seen a near exponential growth in research activity surrounding AI explainability \cite{BARREDOARRIETA202082}.
Where interpretability tends to focus on understanding a prediction based on the inherent mechanics of a model~\cite{liu2022graph}, explainability often focuses instead on post-hoc understanding targeting particular audiences \cite{BARREDOARRIETA202082}.
Numerous studies have explored methods like LIME, SHAP, and TreeExplainer to enhance model interpretability by identifying and simplifying influential features and structures within complex models \cite{ribeiro2016why, lundberg2017unified, lundberg2020local2global}.

Prior work has explored the use of prompting large language models to generate explanations to assist in performance on reasoning tasks \cite{wei2022chain, wang2022self, yao2023tree}.
However, while some work discards these explanations as unwanted generation artifacts, there is limited work exploring these explanations in practice.
Most similar to our work, Gavriilidis, et al.\ (2022) explore combining distilled decision trees with Natural Language Explanations \cite{garcia2018explainable} to provide autonomous underwater vehicle (AUV) operators with rule-based natural language explanations \cite{gavriilidis2022towards}.

\subsection{Network Intrusion Detection}
Network Intrusion Detection (NID) systems which aim to detect malicious behavior in real-world network traffic in real time have been extensively studied \cite{mukherjee1994network, paxson1999bro}.
Early NID systems relied on data mining techniques and machine learning approaches due to their strong performance \cite{lee1999data, buczak2015survey}.
More recently, deep learning based approaches have been also been applied to NID \cite{vinayakumar2019deep}.
However, both traditional and deep learning approaches struggle in practice due to their lack of explainability.
Previous work has explored using XAI approaches to assist with explainability.
Local Explanation Method using Nonlinear Approximation (LEMNA) was introduced to help find decision boundaries in security applications such as binary code analysis \cite{guo2018lemna}.
A number of previous works have also created frameworks which apply SHAP to NID \cite{barnard2022robust, wang2020explainable}.

Despite existing work in NID explainability, nearly all prior work aims to provide additional metrics to assist practitioners in understanding the important features or decision boundaries.
Relatively little attention has been paid to generating natural language explanations which do not rely on any prior knowledge of machine learning.
Further, little work has been done to provide background information as to why particular features may be more relevant than others.

\section{Proposed Method}
\label{sec:method}

In this section we introduce both our proposed approach of generating decision tree explanations as well as our proposed framework for human evaluation of decision tree explanations.

\subsection{\approach}

Given a well-trained decision tree, $\mathcal{T}$, and a vector of input features $x$, our goal is to generate a natural language explanation that describes how $\mathcal{T}$ used $x$ to generate a prediction $\hat{y}$.

The set of all nodes in $\mathcal{T}$ can be denoted as $\mathcal{N}$ where each node $n \in \mathcal{N}$ has an associated decision function $d(n)$ that maps input feature vector $x$ to either \textit{True} or \textit{False} for that particular node, even if it is not a leaf node.
In the case of NID, these are often \textit{Benign} and \textit{Threat} instead.
Then, $\mathcal{E}$ is defined as the set of edges in $\mathcal{T}$ where each edge $e \in \mathcal{E}$ connects two nodes in $\mathcal{N}$.
A path $p$ is a sequence of nodes $n_1, n_2, \dots, n_k$ and their associated edges $e_1, e_2, \dots, e_{k-1}$ such that $n_1$ is the root node of $\mathcal{T}$ and $e_i$ is the edge connecting $n_i$ and $n_{i+1}$ where $i \in [1, k-1]$.
The decision function $d(n)$ can then be used to map the last node $n_k \in p$ to  final prediction $\hat{y} \in \{\textit{Benign}, \textit{Threat}\}$.

Generating an explanation in this setting can then be defined as finding a function which generates a natural language explanation given $\mathcal{T}$, $p(x)$, and $d(n)$.

We experiment with a variety of implementations of this function which can largely be split into two categories: \textit{Rule-Based} and \textit{LLM-Based}.

\subsubsection{Rule-Based Explanations}
\label{subsec:rule-based-explanations}
Rule-based decision tree explanations rely on simple text-based templates for generating explanations.
For instance, one can traverse each node in path $p$ then generate a sentence about why that node was traversed instead of its sibling, such as \textit{"Petal Width was greater than 0.8cm, so the right path was chosen, which favors, Iris virginia."}
While Rule-based decision tree explanations are simple to implement, they have many downsides.
Often, they still require some background knowledge of decision trees by the user, which can lead to poor understanding if the user is unfamiliar with machine learning.
Further, they are not able to incorporate background knowledge to explain how the various features and classes may interact with each other.
For example, they are not capable of indicating why \textit{Petal Width} may be important for distinguishing iris flowers.

\subsubsection{LLM-Based Explanations}
\label{subsec:llm-based-explanations}
LLM-based decision tree explanations (LLM-DTE) seek to remedy the issues with Rule-based explanations by abstracting away the underlying mechanics of the decision tree as well as providing background knowledge to explain the features and classes.
To do this, they leverage LLMs which have been autoregressively pretrained on large corpora spanning a wide range of topics.
Similar to Rule-based explanation approaches, templates are used to convert important data from the path and the decision tree into a text format.
Unlike Rule-based explanations though, the filled-in template is provided as a prompt to a LLM which follows the prompt instructions to generate a natural language decision tree explanation.

We explore a variety of different prompt templates in an attempt to provide high-quality, relevant background knowledge for each explanation.
Following the example of classifying Iris flowers, at each node the language model can be prompted to explain why the decision tree favors some classes more than others when a given feature is beyond a particular threshold.
For example, given input features with high \textit{Petal Width} and high \textit{Sepal Width} which the decision tree classified as \textit{Iris-virginica}, the prompted LLM is able to generate an explanation that describes how \textit{Iris-virginica} tends to have longer and wider petals and sepals compared to the other two species.

Further, additional information can be provided to the language model to assist in generating more nuanced explanations.
For instance, by calculating the nodes along the path that have the highest information gain and providing that information in the prompt template, the LLM is able to emphasize key points along the path where certain classes became heavily favored over others.
This is particularly useful for larger decision trees where not all nodes along a path may include information worth mentioning in the explanation.

LLM-based explanations also allow for tuning the explanation to a target user.
If a user has limited knowledge of network intrusion detection, the model can be prompted to generate an explanation tailored to their level of knowledge.

\subsection{Human Evaluation Framework}
\label{subsec:eval-framework}
For evaluation, we propose a new framework to test the understanding of human evaluators.
To do this, we construct a set of counterfactual quiz questions based on $\mathcal{T}$.
Human annotators are then given a natural language explanation and quizzed with the questions to evaluate their understanding.
Different explanations can then be directly compared based on the proportion of questions correctly answered by human evaluators, as the better explanations will lead to better understanding which will in turn lead to more questions correctly answered.

To create the set of quiz questions, we first construct a set of constraints $\mathcal{C}_p$ for path $p$ which contains the ranges of each input feature that would traverse $p$.
The constraints $\mathcal{C}_p$ for input $x$ to follow path $p$ can be defined as:

\[
C_p(x) = \bigcap_{i=1}^{k-1} c_i(x_j), \text{\quad where }
c_i(x_j) = 
\begin{cases} 
x_j \leq t_i & \text{if } e_i \text{ corresponds to } x_j \leq t_i \\
x_j > t_i & \text{if } e_i \text{ corresponds to } x_j > t_i 
\end{cases}
\]
Where $t_i$ is the threshold for $n_i$. The set of all input vectors \( x \) that satisfy all the constraints in \( C_p \) will traverse the path \( p \) in the decision tree.

We then take $\mathcal{C}_p$ to construct a series of counterfactual questions to test a human annotator's understanding of the decision tree inference.
Specifically, we take each constraint from $\mathcal{C}_p$ and generate a counterfactual question about the true value of an input feature $x_i$ and a hypothetical value of the same feature.

As a simple case study, we use the Iris dataset \cite{misc_iris_53} which is comprised of 150 samples each containing 4 features (\textit{Sepal Length}, \textit{Sepal Width}, \textit{Petal Length}, and \textit{Petal Width}) along with a corresponding species label.
Suppose a decision tree is constructed with the and a sample is taken from the Iris dataset.
We first traverse the decision tree with the sample features to construct a path through the tree.
Given the path, suppose we are able to generate constraints for each feature.
Suppose one of the features, \textit{Petal Width} is constrained to be between 0.8cm and 1.75cm, and the actual true input value of that feature for that sample is 0.9cm.
Using this information, we can then generate a series of counterfactual true/false/unsure questions that test an evaluator's understanding of the decision boundary around this feature.
For instance, one question would be "\textit{If the petal width had been significantly smaller (such as 0.5cm) would the outcome have been the same?}"
If the evaluator has a clear understanding of how the decision was made, they would give the correct answer of \textit{False}.
However, if the evaluator does not have a good understanding of how the decision was made, such as when they are given a poor explanation of the decision, they may give an incorrect answer of \textit{True} or \textit{Unsure}.


This approach has numerous advantages.
First and foremost, it measures not whether the actual explanation is correct, but instead whether the evaluator who read it understands the nuances of the inference itself.
Second, this approach does not rely on gold-labeled explanations, as the quiz questions can be automatically generated for any given decision tree, and the quality of explanation is measured simply by the proportion of correct answers an evaluator gives.
Finally, this approach is entirely agnostic to the explanation approach, as it relies solely on the evaluators' ability to interpret explanations and answer questions.
\section{Experiments}
\label{sec:experiment}

In this section, we conduct our experiments to demonstrate the effectiveness of \approach for generating high quality decision tree explanations for network intrusion detection.

\paragraph{Large Language Model}
The large language model we use to generate explanations for our experiments is GPT-4 due to its high quality generation capabilities shown by other long form generation tasks.
Due to their relatively low quality explanations in initial experiments, results of other models such as GPT-3, GPT-3.5, LLaMA, and LLaMA2 are not reported.

\paragraph{Dataset}
For our network intrusion detection dataset, we use NF-BoT \cite{sarhan2021netflow}, an industry standard dataset containing 600k samples.
Each sample contains a wide range of metadata features to assist in intrusion detection including packet source and destination ports, tcp flags, flow duration, source and destination ip addresses, number of packets exchanged, protocol, and others.
Each sample is also labeled as either \textit{Benign} or \textit{Threat}.

\paragraph{Decision Tree}
We train a simple decision tree on the NF-BoT dataset using Scikit-learn \cite{scikit-learn}.
The max tree depth is limited to 4 to limit the complexity of the tree and resulting explanations.
The trained decision tree achieves a test set accuracy of 98.7\%.
It is worth noting that the goal of this paper is not to improve the performance of decision trees, but instead to help generate high quality explanations of their predictions.

\subsection{Human Evaluation}
For evaluation, we use the human evaluation metric proposed in \S\ref{subsec:eval-framework}.
More specifically, we randomly select 10 samples from the test set, which are given as input to the decision tree.
For each decision tree inference, we programmatically generate a series of quiz questions to test a human evaluator's understanding of the decision tree prediction.
Varying numbers of questions are asked depending on the path the decision tree.
The human evaluators are automatically scored based on whether they correctly identify when the counterfactual inputs would have resulted in different outcomes or not.
The individual human evaluation score is then found by calculating the number of quiz questions correctly answered for a given explanation, and the overall score for the explanation is given by the mean and standard deviation across the evaluators' scores.

Each evaluator is given the same set of questions.
We use 10 evaluators in total.
Half of the evaluators are selected for their background in machine learning while the other half have a background in security.

\subsection{Explanation Approaches}
We experiment with a wide variety of different implementations to generate decision tree explanations which can largely be split into two categories: \textit{Rule-Based} and \textit{LLM-Based}.

\paragraph{Rule Based Explanations}
Following \S\ref{subsec:rule-based-explanations}, we implement a rule-based explanation approach which simply traverses each node in path $p$ then constructs a sentence about why that node was traversed instead of its sibling.
Details about what the individual features represent are provided.

\paragraph{LLM-Based Explanations}
Following \S\ref{subsec:llm-based-explanations}, we also implement an LLM-based explanation approach.
To construct our LLM prompt, we first describe the task of network intrusion detection and provide feature descriptions along with a string-based representation of the trained decision tree.
Similar to our rule-based explanation, we add a sentence for each node traversed in path $p$, providing data such as the feature being considered, the threshold of the node, the split between classes in the training set, and the value of the feature in question.
The predicted label from the decision tree is provided and emphasized.
Finally, the prompt is ended with instructions to describe in simple terms why the decision tree came to its conclusion.

After providing the prompt to the LLM, the generated explanation is recorded.

\subsection{Results}

\begin{table}
  \caption{Human Evaluators' Quiz Scores}
  \label{human-eval-scores}
  \centering
  \begin{adjustbox}{width=0.33\textwidth}
  \begin{tabular}{lll}
    \toprule
    Explanation Approach    &  Quiz Score \\
    \midrule
    Rule-based     & 12.4 $\pm$ 6.5 (out of 25)      \\
    \approach (Ours) & \textbf{17.3 $\pm$ 5.6} (out of 25)     \\
    \midrule
    Overall     & 30.4 $\pm$ 12.1 (out of 50)   \\
    \bottomrule
  \end{tabular}
  \end{adjustbox}
\end{table}

The results from the human evaluators' quiz scores are shown in Table \ref{human-eval-scores}.
The evaluators scored much higher on the questions when given an LLM-based explanation than when given a rule-based explanation, indicating that the LLM-based explanation approach significantly outperforms the rule-based approach.
Specifically, LLM explanations led to 39.5\% higher quiz scores than rule-based approaches.

\begin{table}
  \caption{Human Evaluators' Qualitative Preferences of Rule-based and LLM-based explanations. Results are presented as percentages averaged over evaluators and questions in each category.}
  \label{human-pref-scores}
  \centering
  \begin{adjustbox}{width=1.00\textwidth}
  \begin{tabular}{l|lll|lll|lll}
    \toprule
    \multicolumn{1}{c}{} & \multicolumn{3}{c}{Readability} & \multicolumn{3}{c}{Quality} & \multicolumn{3}{c}{Background Knowledge} \\
    \midrule
    Approach    &  Low & Medium & High & Low & Medium & High & Low & Medium & High  \\
    \midrule
    Rule-based     & 30 $\pm$ 12.2 & 40 $\pm$ 7.1 & 30 $\pm$ 10  & 22 $\pm$ 4.5 & 48 $\pm$ 4.5 & 30 $\pm$ 7.1 & 16 $\pm$ 5.5 & 54 $\pm$ 13.4 & 30 $\pm$ 10  \\
    \approach (Ours) &  12 $\pm$ 10.9 & 46 $\pm$ 11.4 & \textbf{42 $\pm$ 4.5} & 6 $\pm$ 5.5 & 42 $\pm$ 16.4 & \textbf{52 $\pm$ 13.0} & 2 $\pm$ 4.5 & 46 $\pm$ 11.4 & \textbf{52 $\pm$ 13.0}    \\
    \bottomrule
  \end{tabular}
  \end{adjustbox}
\end{table}

Human evaluators were also tasked with qualitatively scoring explanations based on their readability, overall quality, and use of background knowledge on a scale of \textit{Low}, \textit{Medium}, and \textit{High}.
The qualitative preferences of human evaluators are shown in in Table \ref{human-pref-scores}.
Percentages are averaged over all questions.

Evaluators preferred LLM-based explanations to Rule-based ones for readability, quality, and background knowledge.
Specifically, LLM-based explanations were rated highly on readability 15\% more often than rule-based explanations.
Similarly, LLM-based explanations were rated highly nearly twice as often for quality, and 70\% more often for use of background knowledge.

These results indicate not only that human evaluators strongly preferred LLM-based explanations to rule-based ones, but also that our proposed human evaluation framework correlates strongly with human evaluators' preferences.

\section{Conclusion and Future Work}
\label{sec:conclusion}

In this paper, we explore whether large language models can be used to generate explanations for decision trees used for network intrusion detection.
We find LLM generated explanations of decision trees are rated highly by human evaluators in terms of readability, quality, and use of background knowledge.
Further, we introduce a new framework for human evaluation of decision tree explanations that relies on automatically generated quiz questions to test the understanding of human evaluators.

There are a number of exciting directions for future work.
While effort has been put into generating decision tree explanations in the closed book setting, it remains to be seen whether retrieving relevant documents similar to dense passage retrieval (DPR) \cite{karpukhin2020dense} would further improve explanations.
\section*{Limitations}
\label{sec:limitations}
There are a number of limitations of LLM-based decision tree explanations.
Due to their usage of large language models and the length of text being generated, they are relatively slow and expensive to generate.
Further, as the decision tree depth increases, the prompt provided to the language model lengthens and explanation quality begins to degrade.


\clearpage
\section*{Acknowledgements}
This work was supported by ONR N00014-22-1-2507.

\bibliography{bibliography}
\bibliographystyle{unsrt}

\end{document}